# Are Artificial Dendrites useful in NeuroEvolution?

Larry Bull

Computer Science Research Centre

University of the West of England, Bristol BS16 1QY UK

Larry.Bull@uwe.ac.uk

**Abstract**

The significant role of dendritic processing within neuronal networks has become increasingly clear. This letter explores the effects of including a simple dendrite-inspired mechanism into neuroevolution. The phenomenon of separate dendrite activation thresholds on connections is allowed to emerge under an evolutionary process. It is shown how such processing can be positively selected for, particularly for connections between the hidden and output layer, and increases performance.

**Introduction**

Neurons typically receive signals from synapses via dendrites and send signals down their axon to other synapses. It has long been established that the branches of dendritic trees can act as separate subunits with individual activation processing capabilities before their overall activity is integrated by the cell's soma and passed on (e.g., [Koch et al., 1982]). This has led to a number of computational models of their behaviour (e.g., see [Poirazi & Papoutsi, 2020] for a recent review). Dendrite-like processing has previously been included in a very few neuroevolution systems (e.g., see [Wang et al., 2020] for a recent example), as have other forms of extra information processing beyond traditional models, such as synapses (e.g., [Howard et al., 2014]) and gaseous neurotransmitters (e.g., [Philippides et al., 2005]).

To the author's knowledge none of the prior work on dendrites has examined what, if any, benefit the additional mechanism(s) afforded over equivalent traditional neural networks. The use of simulated evolution to enable simple dendrites to emerge between connections within a neural network during the design process is explored here. It is shown how the added computational mechanism can be preferentially exploited between the hidden and output layers on - more complex versions of - a tuneable regression task.

**Tuneable Data from the NK Model**

The well-known NK model [Kauffman & Levin, 1987] is used here to provide a flexible source of data for a regression task. In the NK model the features of the genome/system are specified by two parameters: $N$, the length of the genome; and $K$, the number of genes that has an effect on the fitness contribution of each (binary) gene. Thus increasing $K$ with respect to $N$ increases the epistatic linkage, increasing the ruggedness of the fitness/problem landscape. The increase in epistasis increases the number of optima, increases the steepness of their sides, and decreases their correlation. The model assumes all intragenome interactions are so complex that it is only appropriate to assign random values to their effects on fitness. Therefore for each of the possible $K$ interactions a table of $2^{(K+1)}$ fitnesses is created for each gene with all entries in the range 0.0 to 1.0, such that there is one fitness for each combination of traits (Figure 1). The fitness contribution of each gene is found from its table. These fitnesses are then summed and normalized by $N$ to give the selective fitness of the total genome.

The NK model can be used as a flexible regression function: by altering $K$, the degree of interdependence between the features (genes) in the binary input space is systematically tuneable, as is the number of features by altering $N$.

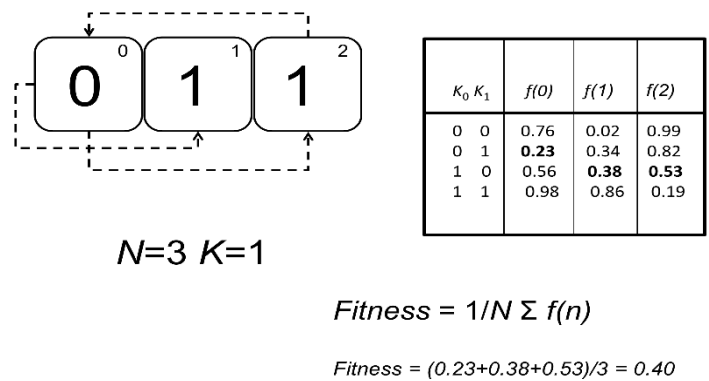

Figure 1: An example NK model showing how the fitness contribution of each gene depends on $K$ random genes. Each gene of the genome has a fitness table created for it (centre gene shown).

**Dendrites as Activation Thresholds: Conditional Connectivity**

Standard multilayered perceptrons (MLP) [Rosenblatt, 1961] are evolved here, with each individual representing the weights for a fully-connected, two-layered network. Each MLP has $N$ input nodes, $H$ hidden layer nodes, and one output node, with all nodes containing a bias and using a sigmoid transfer function (Figure 2, left). The steady-state evolutionary algorithm consists of a population of $P$ initially randomly generated MLPs, uses binary tournament selection of size two for reproduction and random selection for replacement, and gene mutation in a single offspring. Weights are initially seeded in the range [-1.0,1.0] and upon mutation the offspring has one randomly chosen weight adjusted by a random amount from the range [-$R$,+$R$].

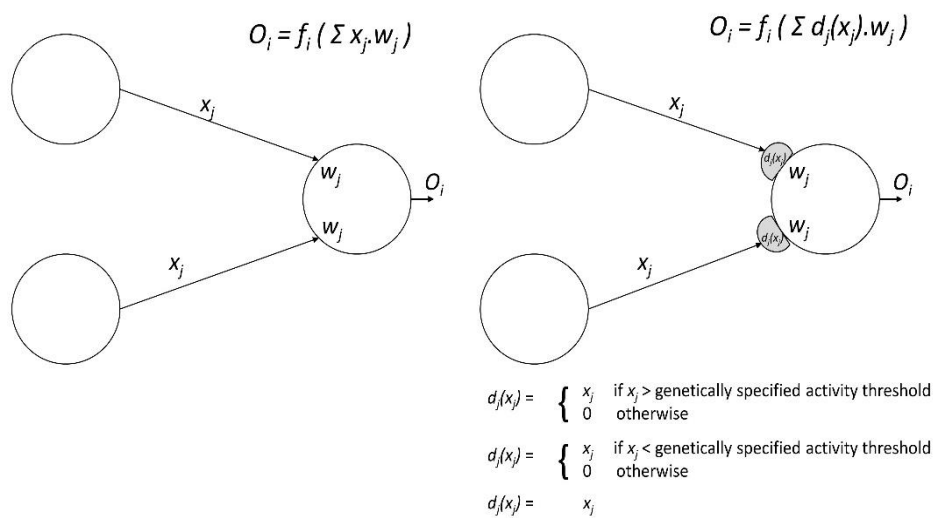

Figure 2: Showing on the left a traditional perceptron, with a transfer function ($f$) using the sum of the weights ($w$) multiplied by the current inputs ($x$) to produce the output ($o$). Showing on the right the additional dendrite processing ($d$) implemented here as one of two simple threshold functions, or as in the traditional model if evolution has not specified the mechanism for the connection.

A single dendritic branch can connect with many synapses and potentially exhibit a variety of computational processes, such as Boolean logic, low pass filtering, and coincidence detection (see [Poirazi & Papoutsi, 2020]). A greatly simplified scheme is adopted here of one synapse connection per branch and an associated activation level discriminator. For example, the excitatory effect of a connection on a dendritic branch can be proportional to its distance from the soma. Here each connection between any two nodes has a binary flag to indicate whether or not it includes dendritic processing. If so, a further binary flag indicates whether the activation threshold is a lower or upper bound, as well as a value for the threshold. If a connection is using such dendritic processing, the current input value is multiplied by its corresponding weight if and only if the threshold criterion is met, otherwise the connection contributes nothing to the activation sum of the node for the current forward pass (Figure 2, right). The binary flags are seeded as inactive and thresholds are seeded randomly in the range [-1.0,1.0]. The mutation process is altered such that either a connection weight is chosen for alteration or a dendrite is chosen. In the latter case, if a connection is not currently using dendritic processing, a random threshold is assigned, along with a value for the flag indicating whether it is an upper or lower bound. If a connection is currently using dendritic processing, either the threshold is altered as weights are, the flag for it being a lower or upper bound is flipped, or the dendrite is disabled. Upon replacement, if the offspring has the same fitness as the selected individual, the MLP using the fewest dendritic connections is chosen. Ties are broken at random. In this way *there is selective pressure against the use of dendrites in the networks*.

All results reported here are the average of twenty runs. A training set of 1000 randomly created NK examples is used per fitness evaluation (mean squared error) and a test set of the same size is used. For each binary genome/data point and its fitness, instead of the typical use of -1.0 to replace zeroes, a zero gene is randomly assigned a value from the range [-1.0,0.0) and a one gene from the range [0.0,1.0] to serve as the feature value to be fed into the network. Here *P*=50, *H*=10, and *R*=0.1.

As Figure 3 shows, standard MLPs can be evolved to reduce the mean squared error to 1% or less on the test set, with *K*=0 being easier than *K*=15, as might be expected. It can also be seen that the MLPs able to exploit the simple dendrite activation thresholds give significantly better test errors (T-test, *p*<0.05) in both cases.

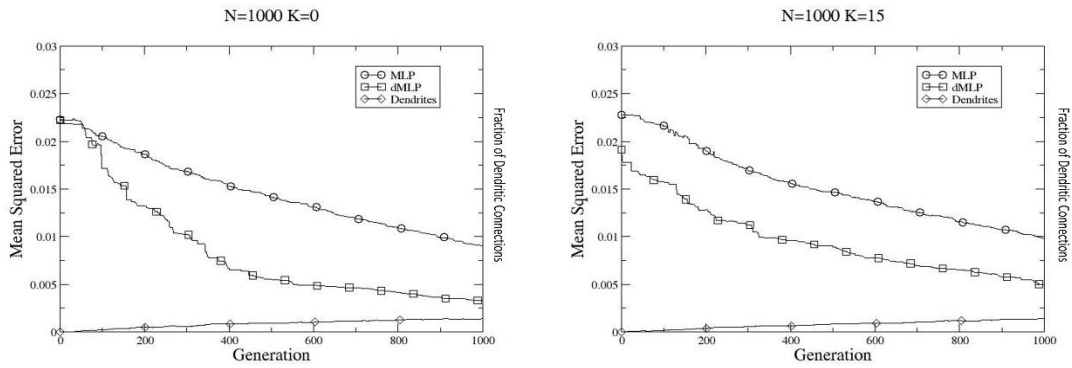

Figure 3: Showing the test error of the best solution, over 1000 generations, on datasets with varying feature interdependence ($K$), with $N$=1000, by networks with (dMLP) and without (MLP) simple dendritic processing enabled. The fraction of connections exploiting dendrites is also shown.

Figure 3 also shows how the fraction of connections exploiting separate dendrites increases over time, typically to around 15 of the total 10,021 connections. Given this is a relatively small number, comparisons were made with a version of the mechanism wherein rather than a dendrite connection using the threshold, the decision to include the current input in the activation sum was made at random on each forward pass. Figure 4 shows how the performance of this mechanism is significantly worse (T-test, $p$<0.05) than both the dendrite threshold MLPs and standard MLPs.

Analysis of where the dendrite processing emerges indicates a tendency towards each output layer node containing one such connection to almost every hidden layer node, with some connections between the hidden layer and the inputs also seen. Results (not shown) from disabling the output layer connections utilising dendrites gives the same performance as the standard MLPs.

Finally, Figure 5 shows how the benefit of the extra mechanism varies with the number of features in the data set. As can be seen, for smaller $N$, no significant difference in training or test error is achieved with the other parameters used here (T-test, $p \geq 0.05$) but as the complexity of the task increases, so a benefit from the simplified dendritic processing arises.

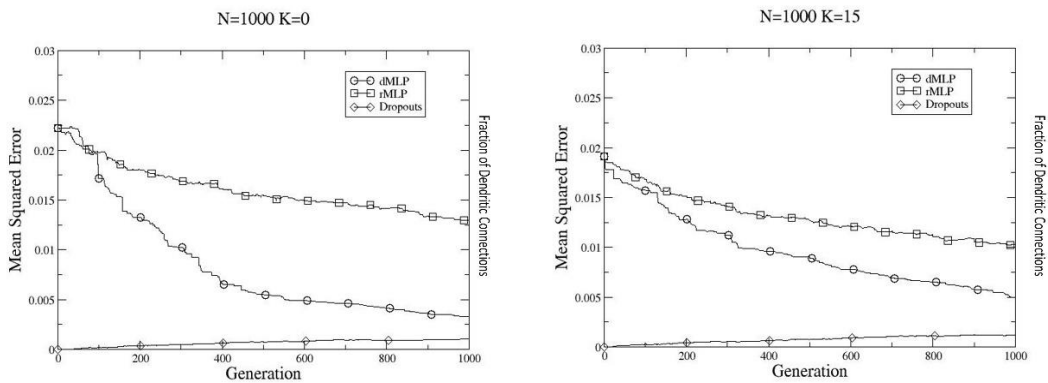

Figure 4: Showing the test errors of the best solution, over 1000 generations, on datasets with varying feature interdependence (*K*), with *N*=1000, by networks evolved with simple dendritic processing (dMLP) and those using random dropout (rMLP). The fraction of connections exploiting dropout is also shown.

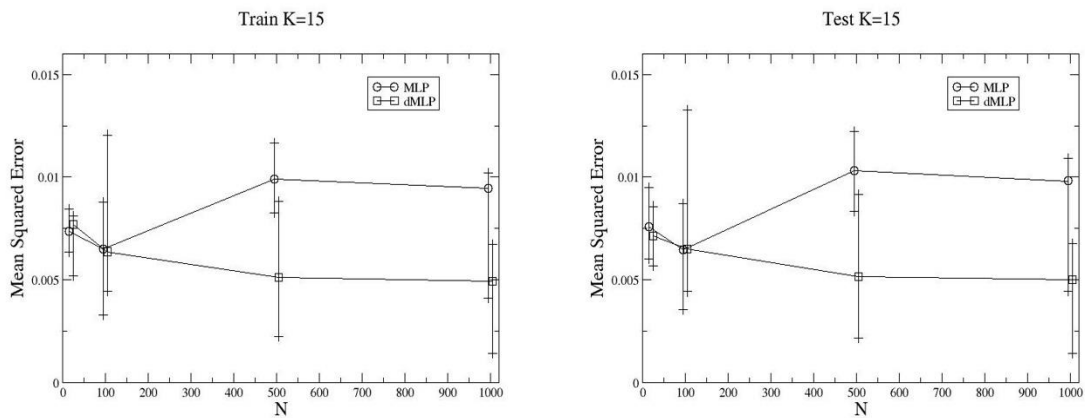

Figure 5: Showing the train and test errors of the best solution reached after 1000 generations, on datasets with significant feature interdependence and a varying number of features (*N*), by networks evolved with (dMLP) and without (MLP) simple dendritic processing enabled. Error bars show min and max values.

**Conclusion**

It is well-established that biological neurons are able to exhibit far more complex computational processes than those assumed in standard artificial neural networks. In particular, post-synaptic processing by dendrites is a significant source of such capabilities. Whilst a very few neuroevolution systems have included dendrites in some form, this is the first time a comparative benefit has been shown. That is, it has been shown that even a very simple dendrite-inspired mechanism can be beneficial – regardless of a selective pressure against its inclusion – so long as the task is sufficiently complex. The simple threshold used was found most beneficial between the hidden and output layers, i.e., not between the input and hidden layers, despite the binary nature of the input space. Results are not significantly changed without the selective pressure against dendrites (not shown).

Given that dendrites may exhibit significant computational capabilities, numerous extensions to the threshold mechanism can be envisaged, particularly with less computational overhead than most other multi-compartmental/network-in-network dendrite models. A simple extension in complexity to a threshold is the specification of a range of input values over which a connection considers an input signal appropriate to be processed. Results (not shown) indicate this slightly more complex variant will also be selected for and is beneficial over the threshold mechanism when, rather than the binary data features being encoded in two non-overlapping ranges as above, a binary one input is encoded as a random sample from the range [-0.5, 0.5] and a zero from either [-1.0,-0.5) or (0.5,1.0]. Dendrite function aside, another obvious avenue of future work is to allow more than one synapse per dendritic branch, i.e., as typically seen in nature.